\documentclass[preprint,12pt]{elsarticle}

\usepackage{amssymb}
\usepackage{amsmath}
\usepackage{subcaption}
\usepackage{soul}
\journal{Nuclear Physics B}

\begin{document}

\begin{frontmatter}



\title{FaceFilterSense: A Filter-Resistant Face Recognition and Facial Attribute Analysis Framework}


\author[inst1]{Shubham Tiwari}
\author[inst1]{Yash Sethia}
\author[inst1]{Ritesh Kumar}
\author[inst1]{Ashwani Tanwar}
\author[inst1]{Rudresh Dwivedi}

\affiliation[inst1]{organization={Department of Computer Science and Engineering},%Department and Organization
            addressline={Netaji Subhas University of Technology (NSUT)}, 
            city={New Delhi},
            postcode={110078}, 
            state={Delhi},
            country={India}}

% \author[inst1,inst2]{Author Three}

% \affiliation[inst2]{organization={Department Two},%Department and Organization
%             addressline={Address Two}, 
%             city={City Two},
%             postcode={22222}, 
%             state={State Two},
%             country={Country Two}}

\begin{abstract}
With the advent of social media, fun selfie filters have come into tremendous mainstream use affecting the functioning of facial biometric systems as well as image recognition systems. These filters vary from beautification filters and Augmented Reality (AR)-based filters to filters that modify facial landmarks. Hence, there is a need to assess the impact of such filters on the performance of existing face recognition systems. The limitation associated with existing solutions is that these solutions focus more on the beautification filters. However, the current AR-based filters and filters which distort facial key points are in vogue recently and make the faces highly unrecognizable even to the naked eye. Also, the filters considered are mostly obsolete with limited variations. To mitigate these limitations, we aim to perform a holistic impact analysis of the latest filters and propose an user recognition model with the filtered images. We have utilized a benchmark dataset for baseline images, and applied the latest filters over them to generate a beautified/filtered dataset. Next, we have introduced a model FaceFilterNet for beautified user recognition. In this framework, we also utilize our model to comment on various attributes of the person including age, gender, and ethnicity. In addition, we have also presented a filter-wise impact analysis on face recognition, age estimation, gender, and ethnicity prediction. The proposed method affirms the efficacy of our dataset with an accuracy of 87.25\% and \hl{an optimal accuracy for facial attribute analysis.}
\end{abstract}

\begin{graphicalabstract}
\includegraphics[scale=0.55]{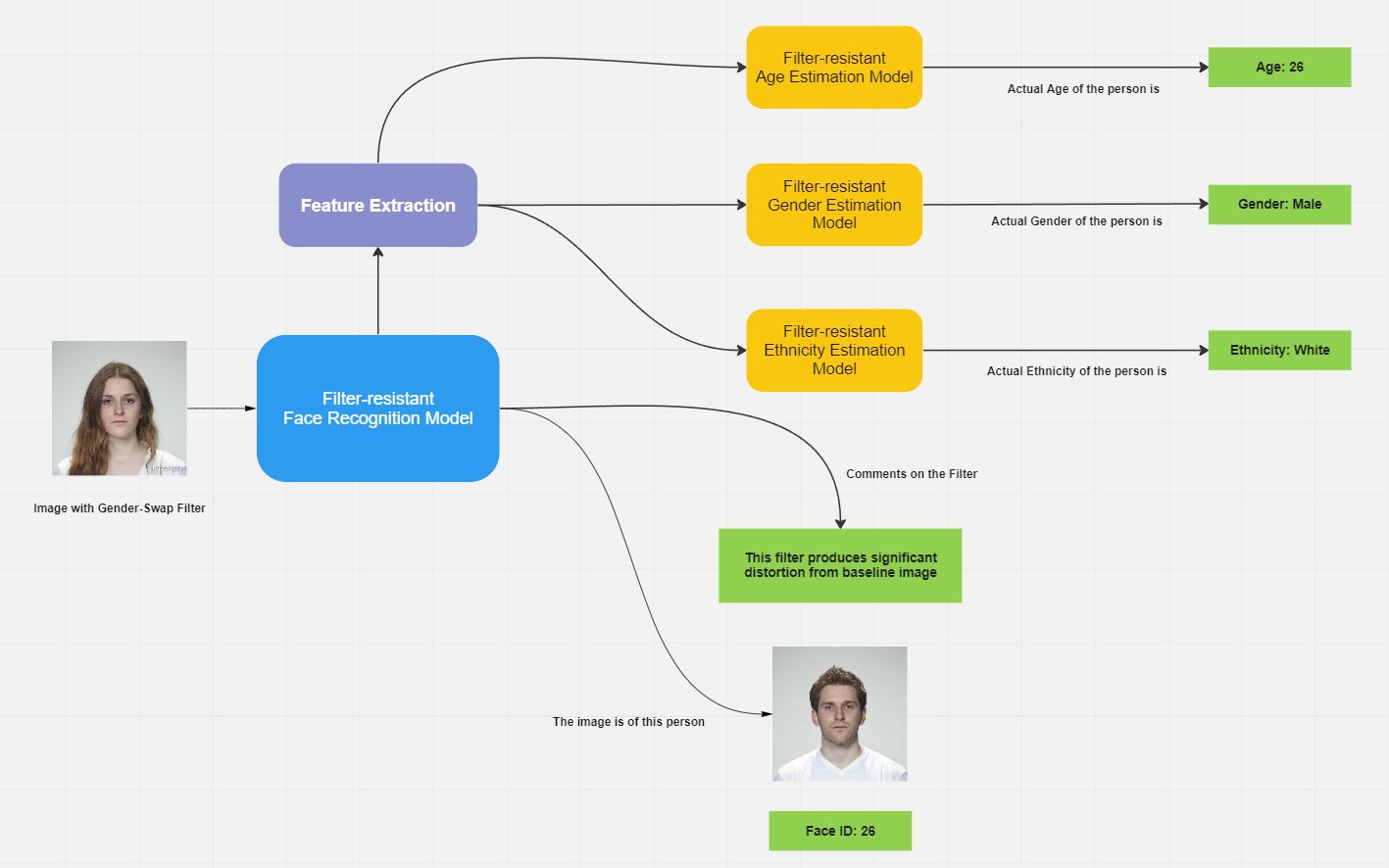}
\end{graphicalabstract}

\begin{highlights}
\item Development of \textbf{FaceFilterNet}: A Filter-Resistant Face Recognition System
\item Development of \textbf{AgeFilterNet}, \textbf{GenderFilter}, and \textbf{EthnicityFilterNet} for the Filter-Resistant analysis of age, gender, and ethnicity from face
\item Filter-wise analysis of distortion produced and comments on the usability of the filter
\end{highlights}

\begin{keyword}
Fun Selfie Filters \sep Biometrics \sep Face Recognition \sep Age estimation \sep Gender Prediction \sep CNN
% keyword one \sep keyword two
%% PACS codes here, in the form: 
\end{keyword}

\end{frontmatter}


\section{Introduction}
\label{A}

\subsection{Background}\label{AA}
Selfies are really popular in today's world of social media. These social media platforms also allow users to add fun selfie filters such as gender swapping, a fake beard or even letting them change their eye colour. These filters focus on beautifying these selfies and making them more interactive. Selfies have now found immense application in biometric identification systems at airports, government IDs, and several other institutions. However, the augmentation of these selfies with custom fun selfie filters has now become a new threat to these face recognition systems.
Such filters impersonate biometric features that may lead to privacy invasion. In this regard, we have earlier created and published a public dataset of facial images with various filters applied. In this work, we are utilizing the dataset to develop a filter-resistant face recognition and facial attribute analysis framework which can accurately identify the person, their age, gender, and ethnicity, with or without filters.

\subsection{Motivations and Contribution}\label{AB}

Few face filters may modify biometric features to a large extent and can promote the intent to impersonate or invade face recognition systems. The authors of Botezatu et al. \cite{botezatu} used fun selfie filters in user authentication in their approach. The authors of Hedman et al. \cite{Hedman_2022} have tried to do similar work on Labelled Faces in the Wild Dataset (LFW). However, the variety of filters used in these approaches were limited and also the performance of existing state-of-the-art methods over filtered images was not up to the mark. To aid in counteracting the above-mentioned issues, we have developed a firstly created and published a public dataset of filtered images for 102 people having 2000+ images. We further developed a facial recognition system which performs well with filtered images as well.

The major contributions of this paper are described below:
\begin{itemize}
	\item We developed a CNN based face recognizer, \textbf{FaceFilterNet}, which recognizes the face even with filters applied.  
	\item We also developed systems for facial attribute analysis which estimates the age, gender and ethnicity of the original face.
	\item We also performed a filter-wise analysis of how much distortion is produced by each filter and comment on the usability of filters.
\end{itemize}

\section{Literature Review}
\label{B}

\begin{table*}[ht!]
\centering
\caption{Comparative Analysis of Related Works.}\label{t_litRev}
\scalebox{0.65}{
\begin{tabular}{ | c | c | c | c | c | c | c | }
\hline
\textbf{Name} & \textbf{Age} & \textbf{Gender} & \textbf{Ethnicity} & \textbf{Beauty Filters} & \textbf{Occlusion Filters} & \textbf{Distortion Filters}
\\ \hline
Botezatu et al. \cite{botezatu} & $ \times $ & $ \times $ & $ \times $ & $ \checkmark $ & $ \checkmark $ & $ \times $
\\ \hline

Hedman et al. \cite{Hedman_2022} & $ \times $ & $ \times $ & $\times$ & $ \checkmark $ & $ \checkmark $ & $ \times $
\\ \hline

Rathgeb et al. \cite{Rathgeb} & $ \times $ & $ \times $ & $ \times $ & $\checkmark$ & $\times $ & $ \times $
\\ \hline

Serengil et al. \cite{serengil2020lightface} \cite{serengil2021lightface} & $ \checkmark $ & $ \checkmark $ & $ \checkmark $ & $ \times $  & $ \times $ & $ \times $
\\ \hline

\textbf{Our Work} & $ \checkmark $ & $ \checkmark $ & $ \checkmark $ & $ \checkmark $ & $ \checkmark $ & $ \checkmark $
\\ \hline

\end{tabular}}
\end{table*}

We explored some previous literature that has worked around face filter analysis and facial retouching. These majorly focused on either beautification filters or filters which introduce large occlusion, but they lacked the usage of the filters which use the facial key points for transforming the face even without much occlusion. Some of these are:
\begin{itemize}
    \item \textbf{Fun Selfie Filters in Face Recognition: Impact Assessment and Removal}: This work by Botezatu et al. \cite{botezatu} very closely discusses the impact of fun selfie filters on facial recognition. However, most of the filters used by the authors are very basic occlusion filters like glasses, hats, bunny ears, etc. These filters introduce severe occlusion in the face which in turn decreases the recognition performance. However, it ignores the impact produced by filters that do not occlude the face but interfere with the facial key points and modify them to produce unidentifiable results. For e.g. Gender reverse or Child. Thus we were interested in including these filters. Also, the paper majorly focuses on the detection and removal of the filter.
    
    \item \textbf{On the Effect of Selfie Beautification Filters on Face Detection and Recognition}: This paper by Hedman et al. \cite{Hedman_2022} uses a dataset of facial images applied with beautification and AR Filters on the standard LFW dataset. The filters applied by them include:
    \begin{itemize}
        \item Dog filter
        \item Glasses filter
        \item Instagram beautification filters like Valencia, Skyline, Ludwig, etc.
        \item Shades with various intensity 
    \end{itemize}

    \begin{figure}
        \centering
        \includegraphics[scale=0.05]{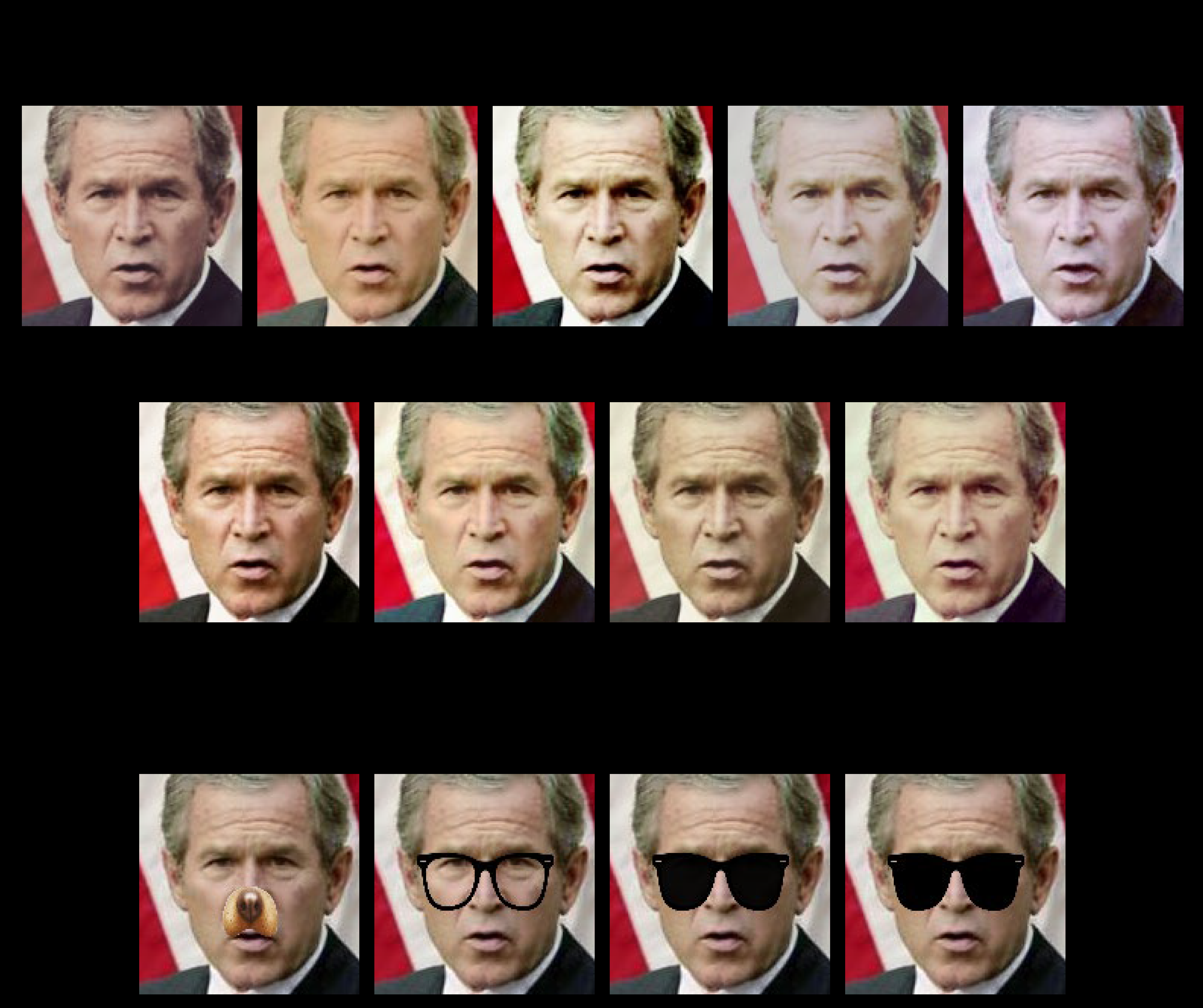}
        \caption{Examples of Filters used by Hedman et al.\cite{Hedman_2022}}
        \label{f3}
    \end{figure}

    Firstly, the AR Filters are very limited in this paper to cover the latest filters. Additionally, the problem with the dataset is that the image size is very small i.e. 64 x 64 due to which the image quality is reduced and further applied filters make the recognition very difficult. Also, the filters used are based on occlusion and are very old. New technology filters are producing very realistic results after the application of filters. Thus, these filters also need to be included.

    \item \textbf{Differential Detection of Facial Retouching: A Multi-Biometric Approach}: In this paper,  Rathgeb et al. \cite{Rathgeb} focused on the facial retouching techniques used by the filters in commonly used mobile applications like BeautyPlus, FotoRus, InstaBeauty etc. They have used multiple databases to test their results using models like COTS, SIFT, PRNU analysis, etc. The focus is majorly on the detection of the filter and filter-wise analysis, however, most of these apps provide beautification filters only.

    \item \textbf{LightFace: A Hybrid Deep Face Recognition Framework}: This work by Serengil et al. \cite{serengil2020lightface} provides a hybrid face recognition framework wrapping state-of-the-art models: VGG-Face, Google FaceNet, OpenFace, Facebook DeepFace, DeepID, ArcFace, Dlib, and SFace. They claim they have surpassed the human-level accuracy of \textbf{97.53\%} on facial recognition tasks. However facial filters are not considered in this work.

    \item \textbf{HyperExtended LightFace: A Facial Attribute Analysis Framework}: This is another work by Serengil et al. \cite{serengil2021lightface} provides a facial attribute analysis (age, gender, emotion, and race) framework for Python. They have analyzed how these parameters are related to facial features. However, they have also not worked around facial filters.
\end{itemize}

\section{Proposed Work}
\label{C}
The proposed work aims to develop a CNN-based filter-resistant face recognition system, perform the facial attribute analysis of features like age, gender, and ethnicity, and perform a filter-wise analysis of the distortion produced by filters to comment on their usability. First, we use the FaceFilterNet model to classify the self-generated filtered facial images (FRLL-Beautified) of 102 persons each with 10 different filters. Next, we used this model as a feature extractor for the estimation of age, gender, and ethnicity. Lastly, we compared the amount of distortion produced by each filter using our custom-defined metric and on the basis of it, commented on the usability of these filters.

\subsection{FaceFilterNet: The filter-resistant face recognizer}\label{CA}
\begin{figure}
        \centering
        \includegraphics[scale=0.25]{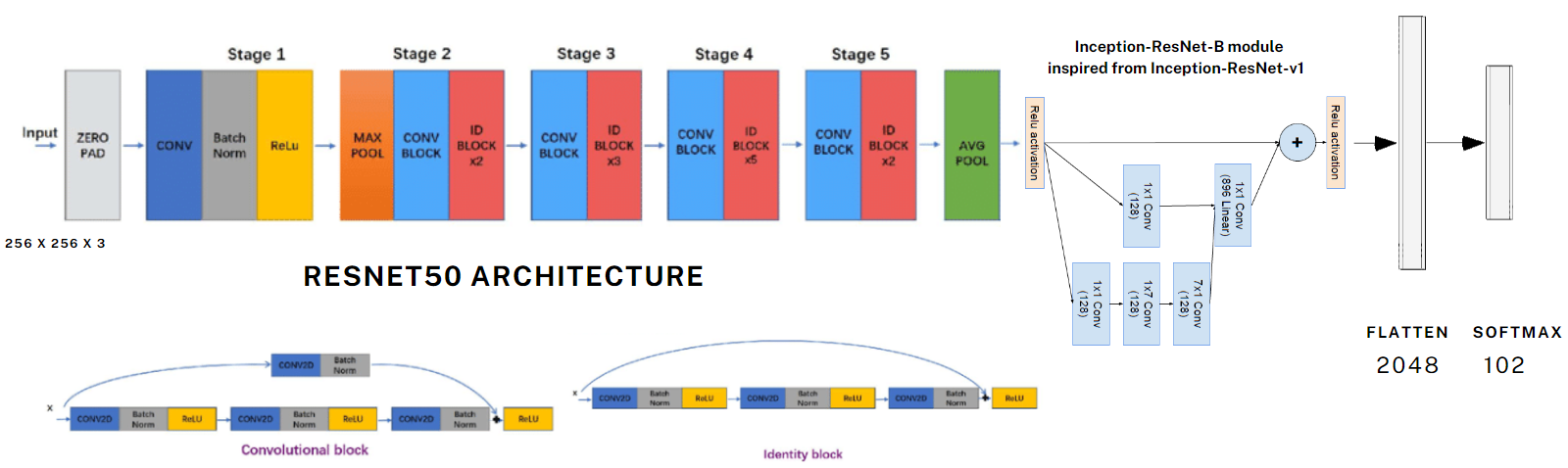}
        \caption{Model Architecture of \textbf{FaceFilterNet} which was further used as feature extractor for facial attribute analysis of age, gender and ethnicity}
        \label{f10}
    \end{figure}

We utilized the FRLL-Beautified dataset for training having facial images of 102 people from 10 profiles and 10 different filters applied to each base image. We split this dataset into a train-test ratio of 80:20. Next we selected an appropriate model architecture. We used the pre-trained architecture of ResNet-50 and added the Inception-ResNet-B module inspired taken from Inception-ResNet-v1 architecture, followed by a Flatten layer of 2048 neurons and finally the output softmax layer of 102 neurons. The model takes the image as an input vector of size 256 x 256 x 3 and returns a vector of size 1 x 102 which gives the probability of the image belonging to the particular class. We call this network as FaceFilterNet and its architecture is shown in Fig. \ref{f10}

\subsection{AgeFilterNet: The filter-resistant age estimator}\label{CB}
    We observed that our face recognizer model, \textbf{FaceFilterNet} has captured the facial features in a way that is resistant to filters. So we can use it as a feature extractor and perform facial attribute analysis for estimating age, gender, and ethnicity. This also helped us speed up the training and validation process because the trained weights of FaceFilterNet provided a good representation of the input. We used this feature extractor and further used dense layers to give output for various facial attributes.

    We analyzed that age prediction is more of a regression problem because age is a continuous random variable. Thus we designed a regression model, \textbf{AgeFilterNet}, for our custom age predictor. The architecture of AgeFilterNet is shown in figure \ref{f11}. 

\subsection{GenderFilterNet: The filter-resistant gender predictor}\label{CC}
\begin{figure}[!t]%
	\centering
	\subfloat[]{\includegraphics[width=0.5\textwidth]{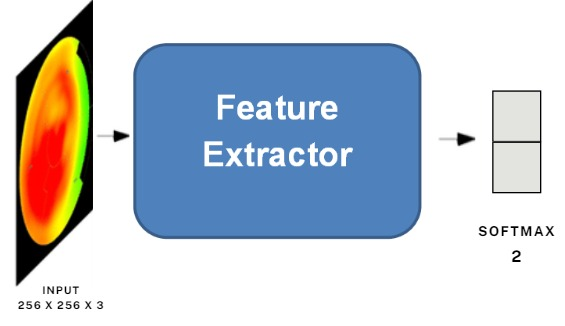}\label{f11}}
	\subfloat[]{\includegraphics[width=0.5\textwidth]{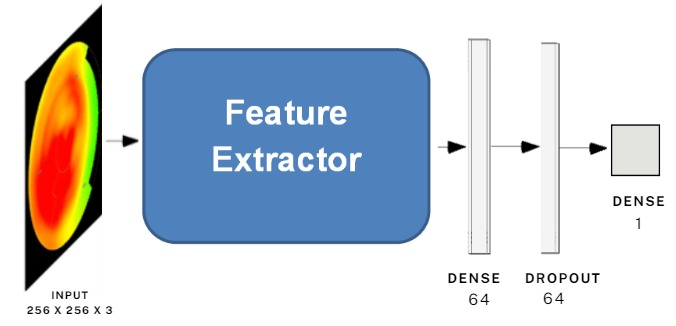}\label{f12}}\\
	\subfloat[]{\includegraphics[width=0.6\textwidth]{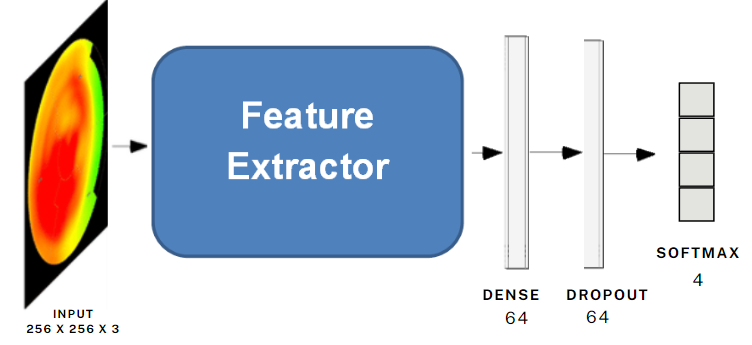}\label{f18}}%
	\caption{(a), (b), and (c) represents the model architecture of GenderFilterNet, AgeFilterNet, and EthnicityFilterNet, respectively}%
	\label{some-label}%
\end{figure}

    Our custom gender model, \textbf{GenderFilterNet}, also utilizes the feature extractor shown in figure \ref{f10}. It is a classification model with 'male' and 'female' as output classes. This is because our dataset only had binary genders. We can expand it to non-binary genders as well with a different dataset. The architecture of GenderFilterNet is shown in figure \ref{f12}.

% \begin{figure*}
%     % \hspace{10pt}
%      \begin{subfigure}[htbp]{0.3\textwidth}
%      \centering
%          \includegraphics[scale=0.3]{images/genderfilternet.png}
%          \caption{GenderFilterNet}
%          \label{f11}
%      \end{subfigure}
%      \hspace{60pt}
%      \begin{subfigure}[htbp]{0.3\textwidth}
%      \centering
%          \includegraphics[scale=0.3]{images/agefilternet.png}
%          \caption{AgeFilterNet}
%          \label{f12}
%      \end{subfigure}
%      \begin{subfigure}[htbp]{0.3\textwidth}
%      \centering
%          \includegraphics[scale=0.3]{images/ethfilternet.png}
%          \caption{EthnicityFilterNet}
%          \label{f18}
%      \end{subfigure}
%     \caption{Model Architectures of our custom age and gender predictors}
%     \label{f_arch}
% \end{figure*}

\subsection{EthnicityFilterNet: The filter-resistant ethnicity predictor}\label{CD}
    Using the same feature extractor, our custom ethnicity model, \textbf{EthnicityFilterNet}, is also a classification model with output classes as East Asian, West Asian, Black and White. This is because our dataset only had limited ethnicities. We can expand it to other races as well with a different dataset. The architecture of \textbf{EthnicityFilterNet} is shown in figure \ref{f18}.

\section{Experimental Results and Discussion}\label{D}
\subsection{Dataset used}\label{DA}
\begin{figure*}[!t]
	\centerline{\includegraphics[scale = 1.1]{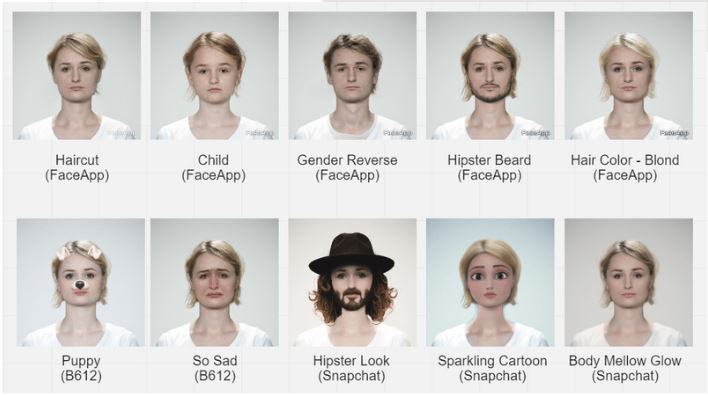}}
	\caption{All the different filters applied to the 'Neutral Front' profile of each person}
	\label{filters}
	\vspace{-0.3cm}
\end{figure*}

Here, we have utilized the well-known FRLL database which contains 1 baseline image and 9 images of the same person in different poses including Neutral Left, Neutral Right, Smiling Left, and Smiling Right. To the neutral front-facing image we apply the different modifications, generating 10 more selfies, each one containing one particular modification. Initially, we had 102 people with 10 poses resulting in 1020 unfiltered images. The application of 10 fun-selfie filters to all 102 images resulted in 1020 more such images. Hence, the total dataset consists of 1020 unfiltered images + 1020 filtered images which results in the generation of 2040 images. This can be shown in fig \ref{filters}.

\subsection{Performance Analysis and Comparison with existing approaches}\label{DB}
The performance analysis of our custom models shows significant improvement over the existing state-of-the-art methods. We have used accuracy, precision, recall, and F-Score as the performance metric for the analysis of the classification models.  While for the regression-based models, we used R2 Score, Mean Absolute Error, and Mean Squared Error as the performance metrics. For face recognition, we are comparing three state-of-the-art methods namely - \textbf{ArcFace\cite{DBLP:arcface}}, \textbf{FaceNet512\cite{facenet}} and \textbf{VGGFace\cite{vggface}}, provided by deepface library \cite{serengil2020lightface}. ArcFace is a deep CNN-based model which is considered to be one of the most effective face recognition systems. Next, FaceNet comes with the inception module architecture which reduces the depth and number of trainable parameters without compromising on the accuracy. VGGFace is a face recognizer that performs significantly well in the wild. Table \ref{t0} shows the performance comparison of FaceFilterNet with these methods. This clearly shows how FaceFilterNet outperforms all these methods in the case of filtered images with an accuracy of \textbf{87.25\%}. This signifies that our model is able to recognize the correct person even after the application of filters.

\begin{table}[ht!]
\centering
    \caption{Performance Comparison of FaceFilterNet with existing state-of-the-art methods}\label{t0}
    \scalebox{0.9}{
        \begin{tabular}{| c | c | c | c | c |}
        \hline
        Performance Metric & ArcFace\cite{DBLP:arcface} & FaceNet512\cite{facenet} & VGGFace\cite{vggface} & \textbf{FaceFilterNet} \\
        \hline
        Accuracy & 0.812 & 0.835 & 0.703 & 0.872 \\
        Precision & 0.790 & 0.834 & 0.702 & 0.879 \\ 
        Recall & 0.789 & 0.829 & 0.706 & 0.875 \\
        F1 Score & 0.809 & 0.841 & 0.701 & 0.871 \\ \hline
        \end{tabular}
    }
\end{table}

Similarly, for the estimation of age, gender, and ethnicities from these filtered images, we used the HyperExtended LightFace \cite{serengil2021lightface} which is a facial attribute analysis framework. It claims to have surpassed human-level accuracy with its deep CNN-based models. The experimental results of the models in this framework are reported in Table \ref{t1}, \ref{t2}, and \ref{t3} for Age, Gender, and Ethnicity respectively. 
% as shown in table \ref{t1} - age
% as shown in table \ref{t2} - gender

As it is clearly evident that the DeepFace Age model produced an average error of \textbf{6-7 years} in age prediction due to these filters. However, AgeFilterNet was able to achieve a much lesser error of around \textbf{one and a half years} on average. This signifies that our model can differentiate between a person who is actually age 60 and someone who is made to look like age 60 using filters. This can be very helpful in the prevention of fake age fraud and other malicious activities related to it. 

\begin{table}[ht!]
\centering
    \caption{Performance Comparison of DeepFace Age Model and AgeFilterNet}\label{t1}
    \scalebox{1}{
        \begin{tabular}{| c | c | c |}
        \hline
        \textbf{Performance Metric} & \textbf{DeepFace Age Model} & \textbf{AgeFilterNet} \\
        \hline
        R2 Score & -0.35 & 0.904 \\ 
        Mean Absolute Error & 6.81 & 1.74 \\
        Mean Squared Error & 67.61 & 4.78\\ \hline
        \end{tabular}
    }
\end{table}

\begin{table}[ht!]
\centering
    \caption{Performance Comparison of DeepFace Gender Model and GenderFilterNet}\label{t2}
    \scalebox{0.9}{
        \begin{tabular}{| c | c | c |}
        \hline
        \textbf{Performance Metric} & \textbf{DeepFace Gender Model} & \textbf{GenderFilterNet} \\
        \hline
        Accuracy & 0.733 & 0.983 \\
        Precision & 0.740 & 0.983 \\ 
        Recall & 0.729 & 0.983 \\
        F1 Score & 0.728 & 0.983\\ \hline
        \end{tabular}
    }
\end{table}

\begin{table}[ht!]
\centering
    \caption{Performance Comparison of DeepFace Ethnicity Model and EthnicityFilterNet}\label{t3}
    \scalebox{0.85}{
        \begin{tabular}{| c | c | c |}
        \hline
        \textbf{Performance Metric} & \textbf{DeepFace Ethnicity Model} & \textbf{EthnicityFilterNet} \\
        \hline
        Accuracy & 0.754 & 0.832 \\
        Precision & 0.437 & 0.772 \\ 
        Recall & 0.303 & 0.715 \\
        F1 Score & 0.305 & 0.710\\ \hline
        \end{tabular}
    }
    
\end{table}

Similarly, in table \ref{t2}, we can see that the gender prediction of DeepFace was accurate up to \textbf{93\%}. Thus filters are responsible for introducing an of \textbf{7\%} in gender prediction. However, in the case of GenderFilterNet, the accuracy is around \textbf{98\%}. This signifies that our model is able to predict the correct gender irrespective of the filter applied. This shows that the model is able to differentiate between a person who is actually a male and someone who is made to look like male using filters. This is also very helpful in detecting online scams majorly through social media. People often fake their identities by using these filters. A lot of fake profiles of women are there on social media which use such filters to fool others. Such acts are often related to crimes and malicious activities. Thus our model can prove to be beneficial in such cases. 

For the ethnicity model, we have used Accuracy, Precision, Recall, and F-Score as the metrics for evaluation as shown in table \ref{t3}. It is clearly evident that the DeepFace ethnicity model increased the precision by \textbf{35\%}. Significant improvements are seen in other metrics as well. This can be very helpful in preventing the ethnicity or racial bias produced by these fun selfie filters.

\subsection{Filter-wise Analysis and discussions on usability}\label{CE}
After the exhaustive analysis of facial recognition and prediction of facial attributes like Age, Gender, and Ethnicity with the help of our custom model, we were quite impressed by the results produced. We then used these results to analyze the effects of filters on facial recognition and the estimation of attributes. Suppose the prediction probability vector for two images is given by X1 and X2 given by equation \ref{eq1} and \ref{eq2} respectively.

\begin{equation} \label{eq1}
X_1 = \begin{bmatrix} x_{11} & x_{12} & ... & x_{1n} \end{bmatrix}
\end{equation}

\begin{equation} \label{eq2}
X_2 = \begin{bmatrix} x_{21} & x_{22} & ... & x_{2n} \end{bmatrix}
\end{equation}

Assume the number of classes in the dataset to be n. The corresponding L2-norms are LX1 and LX2 given by equation \ref{eq3} and \ref{eq4} respectively. 
\begin{equation} \label{eq3}
LX_1 = L_2Norm(X_1) = \frac{X_1}
                      {\sqrt{X_1^T X_1}}
\end{equation}

\begin{equation} \label{eq4}
LX_2 = L_2Norm(X_2) = \frac{X_2}
                      {\sqrt{X_2^T X_2}}
\end{equation}

To understand the impact of filters on facial recognition, we used the metric, \textbf{Average L2 Euclidean Distance, D} given by \ref{eq5}. 
\begin{equation} \label{eq5}
D = \sqrt{(LX_2 - LX_1)^T(LX_2 - LX_1)}
\end{equation}

\subsubsection{Filter-wise Analysis of Face Recognition}\label{CEA} 
We consider a filter to be distorting if the average distance between the baseline image and the image filtered with that particular filter is very high. We calculated this metric for each filter to evaluate the average distortion produced by each filter. The results are shown in Table \ref{f13}.

\begin{table}[]
	\centering
	\begin{tabular}{|l|c|}
		\hline
		\textbf{Filter Name} & \begin{tabular}[c]{@{}l@{}}\textbf{Average L2-Euclidean}\\ \textbf{Distance}\end{tabular} \\ \hline
		Haircut Filter FaceApp & 0.458548\\ \hline
		Child Filter FaceApp  & 0.507977\\ \hline
		Gender Reverse Filter FaceApp & 0.580298\\ \hline
		Hipster Beard Style Filter FaceApp & 0.332914\\ \hline
		Hair Color Blonde Filter FaceApp & 0.222967\\ \hline
		Puppy Filter B612 & 0.494714\\ \hline
		So Sad Filter B612 & 0.191711\\ \hline
		Hipster Look Filter Snapchat & 1.179643\\ \hline
		Sparkling Cartoon Filter Snapchat & 0.615396\\ \hline
		Body Mellow Glow Filter Snapchat & 0.427381\\
		\hline
	\end{tabular}
	\caption{Amount of distortion produced by each filter to the face images in FRLL-Beautified}
	\label{f13}
\end{table}

%\begin{figure}
%    \centering
%    \includegraphics[scale=0.6]{images/df_filter.pdf}
%    \vspace{-60pt}
%    \caption{Amount of distortion produced by each filter to the face images in FRLL-Beautified}
%    \label{f13}
%\end{figure}

As per the standard threshold followed by Serengil et al., \cite{serengil2020lightface} if the value of D given by equation \ref{eq5} exceeds \textbf{0.75}, it is considered to be breaking. We can clearly see from Table \ref{f13} that only the filter - \textbf{Hipster Look} from Snapchat is where our model fails. This can be seen as an indication that the amount of distortion produced by this filter is so high that the face becomes unrecognizable. Similarly, we can see the filters \textbf{Sparkling Cartoon} from Snapchat and \textbf{Gender Reverse} from FaceApp also produce a significantly large amount of distortion.  

\subsubsection{Filter-wise Analysis of Age Estimation}\label{CEB}

%\begin{figure}
%    \centering
%    \includegraphics[scale=0.6]{images/df_age_analysis.pdf}
%    \vspace{-60pt}
%    \caption{Filter-wise age mispredictions. The columns show the average reduction, increment, and net deviation in age estimation due to a given filter.}
%    \label{f15}
%\end{figure}

\begin{table}[]
\resizebox{\textwidth}{!}{%
	\begin{tabular}{|l|c|c|c|}
		\hline
		\multicolumn{1}{|l|}{\textbf{Filter Name}} & \multicolumn{1}{l|}{\begin{tabular}[c]{@{}l@{}}\textbf{Average Age}\\ \textbf{Reduction (yrs)}\end{tabular}} & \multicolumn{1}{l|}{\begin{tabular}[c]{@{}l@{}}\textbf{Average Age}\\ \textbf{Increment (yrs)}\end{tabular}} & \multicolumn{1}{l|}{\textbf{Net Deviation (yrs)}} \\ \hline
		Haircut Filter FaceApp & -2.062500 & 1.478261 & -0.292120\\ \hline
		Child Filter FaceApp & -3.371795 & 1.888889 & -0.741453\\ \hline
		Gender Reverse Filter FaceApp & -3.492754 & 2.000000 & -0.746377\\ \hline
		Hipster Beard Style Filter FaceApp & -2.530303 & 1.812500 & -0.358902\\ \hline
		Hair Color Blonde Filter FaceApp & -2.369863 & 1.636364 & -0.366750\\ \hline
		Puppy Filter B612 & -3.200000 & 1.800000 & -0.700000\\ \hline
		So Sad Filter B612 & -2.142857 & 1.733333 & -0.204762\\ \hline
		Hipster Look Filter Snapchat & -5.774194 & 2.166667 & -1.803763\\ \hline
		Sparkling Cartoon Filter Snapchat & -3.208333 & 3.153846 & -0.027244\\ \hline
		Body Mellow Glow Filter Snapchat & -2.716049 & 1.200000 & -0.758025\\
		\hline
	\end{tabular}}
	\caption{ Filter-wise age mispredictions: The columns show the average reduction, increment, and net deviation in age estimation due to a given filter.}
	\label{f15}
\end{table}

Moving forward we performed the filter-wise analysis of facial attributes. For the analysis of age estimation, we found that some filters made the model overestimate the age while others made it underestimate the age. So we tried to calculate the average reduction and average increment in age estimation for each filter. Suppose the actual age of a person is \begin{math}a_i\end{math} and the predicted age is \begin{math}\hat{a}_i\end{math}, then the average reduction, increment and net deviation in age are calculated using the equations \ref{eq6}, \ref{eq7} and \ref{eq8} respectively.

\begin{equation} \label{eq6}
    \text{Avg age Reduction} = 
    \begin{cases}
        \frac{1}{n_1} \sum_{i=1}^{n_1} (a_i - \hat{a}_i), & \text{if } \hat{a}_i < a_i \text{ for \begin{math}n_1\end{math} instances} \\
        0, & \text{otherwise}
    \end{cases}
\end{equation}

\begin{equation} \label{eq7}
    \text{Avg age Increment} = 
    \begin{cases}
        \frac{1}{n_2} \sum_{i=1}^{n_2} (a_i - \hat{a}_i), & \text{if } \hat{a}_i > a_i \text{ for \begin{math}n_2\end{math} instances} \\
        0, & \text{otherwise}
    \end{cases}
\end{equation}

\begin{equation} \label{eq8}
    \text{Net Deviation} = 
        \frac{1}{N} \sum_{i=1}^{N} (a_i - \hat{a}_i), \text{where } N = n_1 + n_2
\end{equation}

Table \ref{f15} shows the deviation in age estimation by the application of a filter that made the model estimate the age wrongly. It is clearly visible that in general, the model underestimates the age for each filter. This is also evident from the fact that the model produces a mean squared error of \textbf{1.74} years. However, filters like \textbf{Hipster Look} from Snapchat, \textbf{Child Filter} and \textbf{Gender Reverse} from FaceApp tend to make the faces look younger. Similarly, filters like \textbf{Sparkling Cartoon} from Snapchat make the faces look older.

\subsubsection{Filter-wise Analysis of Gender Prediction}\label{CEC}
\begin{figure*}[!htbp]
    \includegraphics[scale=0.3]{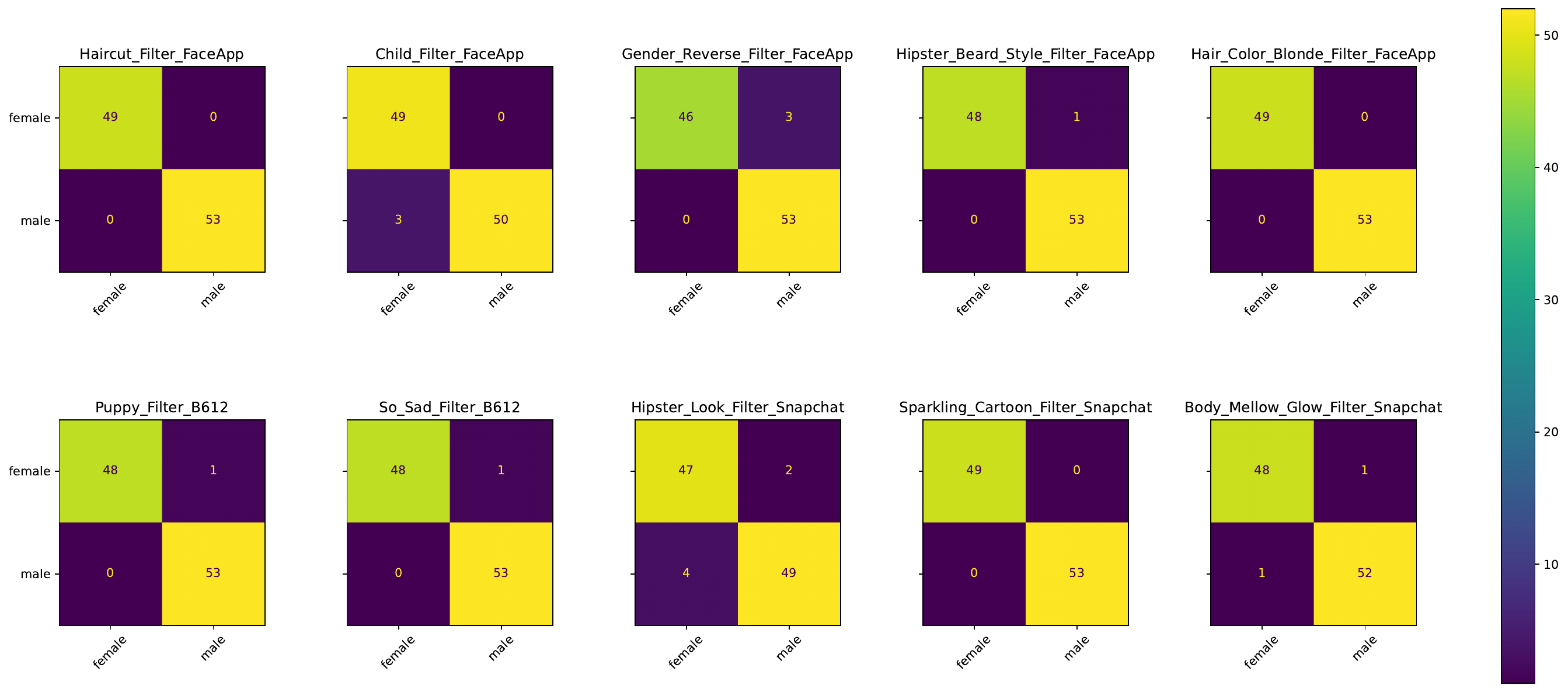}
    \caption{Filter-wise confusion matrices for gender prediction}
    \label{f19}
\end{figure*}

%\begin{figure*}
%    \centering
%    \includegraphics[scale=0.6]{images/df_gender_mispredictions.pdf}
%    \vspace{-60pt}
%    \caption{Filter-wise gender mispredictions: The columns show the number of instances the gender was wrongly predicted for a given filter}
%    \label{f14}
%\end{figure*}

\begin{table}[!t]
	\resizebox{\textwidth}{!}{%
		\begin{tabular}{|l|c|c|}
			\hline
			\textbf{Filter} & \begin{tabular}[c]{@{}c@{}}\textbf{Number of Males} \\\textbf{classified as Females}\end{tabular} & \begin{tabular}[c]{@{}c@{}}\textbf{Number of Females}\\ \textbf{classified as Males}\end{tabular} \\ \hline
			Haircut Filter FaceApp & 1 & 0 \\ \hline
			Child Filter FaceApp & 4 & 0 \\ \hline
			Gender Reverse Filter FaceApp & 1 & 3 \\ \hline
			Hipster Beard Style Filter FaceApp & 0 & 0 \\ \hline
			Hair Color Blonde Filter FaceApp & 1 & 0 \\ \hline
			Puppy Filter B612 & 1 & 1 \\ \hline
			So Sad Filter B612 & 0 & 0 \\ \hline
			Hipster Look Filter Snapchat & 5 & 2 \\ \hline
			Sparkling Cartoon Filter Snapchat & 1 & 0 \\ \hline
			Body Mellow Glow Filter Snapchat & 1 & 0 \\ \hline
		\end{tabular}%
	}
	\caption{Filter-wise gender mispredictions: The columns show the number of instances the gender was wrongly predicted for a given filter}
	\label{f14}
\end{table}

For the analysis of gender prediction, we tried to calculate the number of instances when the model predicted the wrong gender. Figure \ref{f19} shows filter-wise confusion matrices for gender predictions. Table \ref{f14} shows the number of instances when the application of a filter made the model predict the gender wrongly. It is clearly visible that the number of times the model predicted the wrong gender (at max 5) is much lesser than the number of times it predicted correctly (at least 97) for a given filter. So the model is able to capture the artificial features introduced by filters to alter the gender. However, filters like - \textbf{Gender Reverse} from FaceApp produces some images which make them look more like a male. Similarly filters like - \textbf{Hipster Look} from Snapchat and \textbf{Child Filter} from FaceApp produce images that make them look more like a female. Thus these filters might be used maliciously to fake gender.

\subsubsection{Filter-wise Analysis of Ethnicity Prediction}\label{CED}
\begin{figure}[!htbp]
    \includegraphics[scale=0.3]{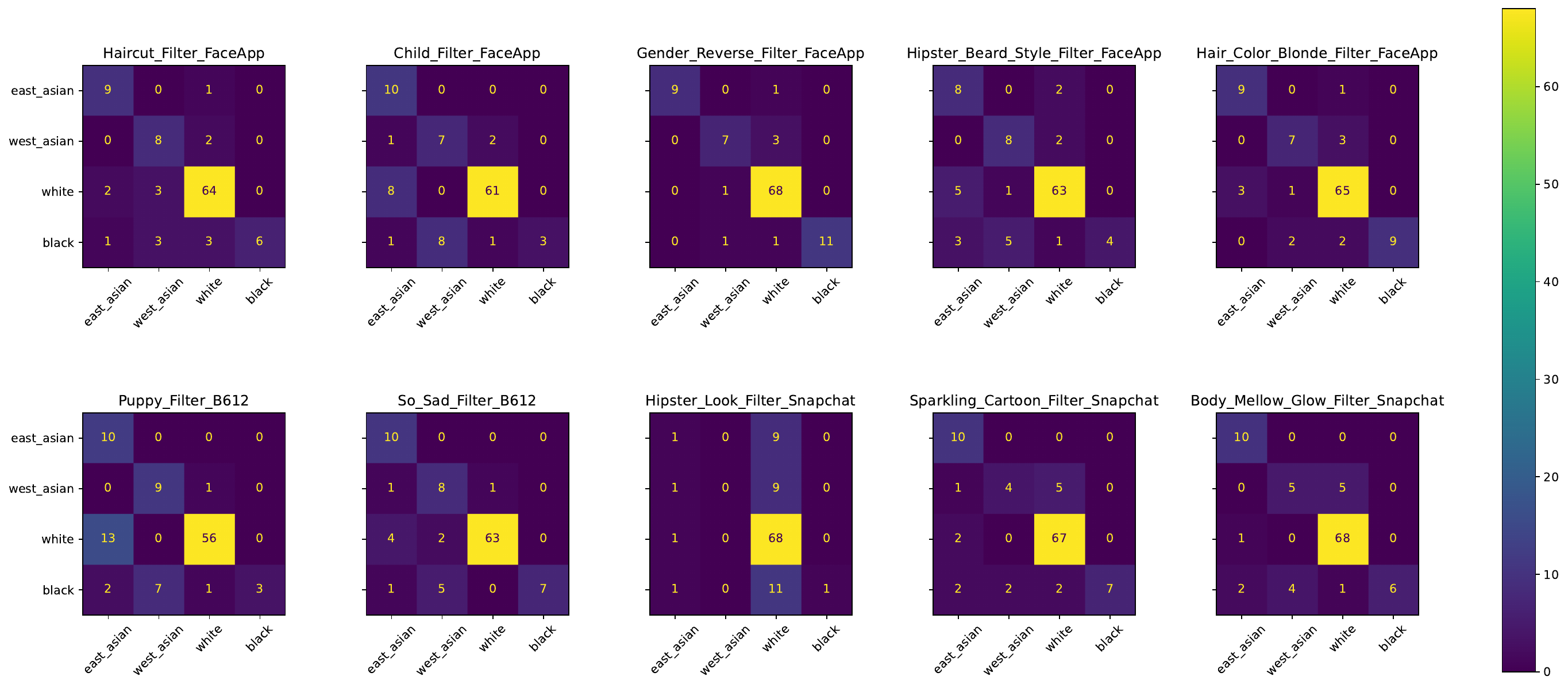}
    \caption{Filter-wise confusion matrices for ethnicity prediction}
    \label{f20}
\end{figure}

%\begin{figure}
%    \centering
%    \includegraphics[scale=0.6]{images/df_eth_mispredictions.pdf}
%    \vspace{-60pt}
%    \caption{Filter-wise ethnicity mispredictions: Each column shows the number of instances when the wrongly predicted class was the given column name}
%    \label{f16}
%\end{figure}

\begin{table}
	\centering
	\resizebox{\textwidth}{!}{%
	\begin{tabular}{|l|l|l|l|l|l}
		\hline
		\textbf{Filter} & \textbf{black} & \textbf{east asian} & \textbf{west asian} & \textbf{white}\\  
		\hline
		Haircut Filter FaceApp & 0 & 3 & 6 & 6\\ \hline
		Child Filter FaceApp & 0 & 10 & 8 & 3\\ \hline
		Gender Reverse Filter FaceApp & 0 & 0 & 2 & 5\\ \hline
		Hipster Beard Style Filter FaceApp & 0 & 8 & 6 & 5\\ \hline
		Hair Color Blonde Filter FaceApp & 0 & 3 & 3 & 6\\ \hline
		Puppy Filter B612 & 0 & 15 & 7 & 2\\ \hline
		So Sad Filter B612 & 0 & 6 & 7 & 1\\ \hline
		Hipster Look Filter Snapchat & 0 & 3 & 0 & 29\\ \hline
		Sparkling Cartoon Filter Snapchat & 0 & 5 & 2 & 7\\ \hline
		Body Mellow Glow Filter Snapchat & 0 & 3 & 4 & 6\\ \hline
	\end{tabular}
	}
	\caption{ Filter-wise ethnicity mispredictions: Each column shows the number of instances when the wrongly predicted class was the given column name.}
	\label{f16}
\end{table}

For the analysis of ethnicity prediction, we tried to calculate the number of instances when the model predicted the wrong ethnicity. Figure \ref{f20} shows the filter-wise confusion matrix for ethnicity prediction. Table \ref{f16} shows the number of instances when the application of a filter made the model predict the ethnicity wrongly. It is clearly visible that filters like - \textbf{Child Filter} from FaceApp and \textbf{Puppy Filter} from B612 misclassify most faces as east asian. This shows that filters like these might promote bias against this ethnicity. Similarly, \textbf{Hipster Look} from Snapchat misclassifies most faces as white. Thus these filters can be marked for scrutiny over ethnicity-based bias.

\subsection{Challenges Faced}\label{DC}
First of all, it was very difficult to find a publicly available dataset that includes facial images with a variety of filters applied. Most of the existing datasets covered only a small number of filters. Thus we had to create a custom dataset for the same. Apart from that, while performing facial attribute analysis, there is a difference in the notion of ethnicities to be considered. Different models define ethnicities in their own way. thus it was a difficult task to compare the ethnicity models. We had to map various names to the ethnicity classes for a holistic comparison. 

\section{Conclusion and Future Scope}
We have successfully assessed the impact of several fun selfie filters on the performance of existing face recognition systems like Dlib, VGGFace + MTCNN, and ArcFace + RetinaFace. We have also proposed a facial recognition model built using ResNet that performs better than the currently available models in the public domain. The model further comments on various attributes of the person including age, gender, and ethnicity. We also published a Python library for the same purpose, bringing all our models and features under a single umbrella. In the future, the model can be trained on publicly available datasets to increase its accuracy, since accuracy is a very stringent measure when dealing with facial recognition systems. Further, the dataset including more varied genders and ethnicities can be used to understand the impact of filters in a better way. Also, this work is not exhaustive work as the field of facial filters is continuously evolving. Hence with the advent of new filters which modify facial features in some other way, those filters need to be included in this type of analysis work. However, we believe that with our work we have laid the foundation for a standard methodology to be followed in the analysis of facial filters.

% \appendix

% \section{Sample Appendix Section}
% \label{sec:sample:appendix}
% Lorem ipsum dolor sit amet, consectetur adipiscing elit, sed do eiusmod tempor section incididunt ut labore et dolore magna aliqua. Ut enim ad minim veniam, quis nostrud exercitation ullamco laboris nisi ut aliquip ex ea commodo consequat. Duis aute irure dolor in reprehenderit in voluptate velit esse cillum dolore eu fugiat nulla pariatur. Excepteur sint occaecat cupidatat non proident, sunt in culpa qui officia deserunt mollit anim id est laborum.

 \bibliographystyle{elsarticle-num} 
 \bibliography{cas-refs}





% \end{thebibliography}
\end{document}